\documentclass{article}
\usepackage{bbding}
\usepackage{multirow}
\usepackage{enumitem}
\usepackage{verbatim}
\usepackage{PRIMEarxiv}
\usepackage{enumitem}
\usepackage{verbatim}
\usepackage{fancyhdr}
\usepackage[utf8]{inputenc} 
\usepackage[T1]{fontenc}    
\usepackage{hyperref}       
\usepackage{url}            
\usepackage{booktabs}       
\usepackage{amsfonts}       
\usepackage{nicefrac}       
\usepackage{microtype}      
\usepackage{lipsum}
\usepackage{fancyhdr}       
\usepackage{graphicx}       
\graphicspath{{media/}}     

\title{On Specifying for Trustworthiness}

\author{
   Dhaminda B. Abeywickrama \Envelope \\ 
   Department of Computer Science \\
   University of Bristol, UK \\
   \texttt{dhaminda.abeywickrama@bristol.ac.uk} 
   \And
   Amel Bennaceur \\
   Department of Computing \\
   Open University, UK \\
   \texttt{amel.bennaceur@open.ac.uk} 
   \And
   Greg Chance \\
   Department of Computer Science \\
   University of Bristol, UK \\
   \texttt{greg.chance@bristol.ac.uk} 
   \And
   Yiannis Demiris \\
   Electrical and Electronic Engineering \\
   Imperial College London, UK \\
   \texttt{y.demiris@imperial.ac.uk} 
   \And
   Anastasia Kordoni \\
   Department of Psychology \\
   Lancaster University, UK \\
   \texttt{a.kordoni@lancaster.ac.uk} 
   \And
   Mark Levine \\
   Department of Psychology \\
   Lancaster University, UK \\
   \texttt{mark.levine@lancaster.ac.uk} 
   \And
   Luke Moffat \\
   Department of Sociology \\
   Lancaster University, UK \\
   \texttt{l.moffat1@lancaster.ac.uk} 
   \And
   Luc Moreau \\
   Department of Informatics \\
   King's College London, UK \\
   \texttt{luc.moreau@kcl.ac.uk} 
   \And
   Mohammad Reza Mousavi \\
   Department of Informatics \\
   King's College London, UK \\
   \texttt{mohammad.mousavi@kcl.ac.uk} 
   \And
   Bashar Nuseibeh \\
   Department of Computing \\
   Open University, UK \\
   \texttt{b.nuseibeh@open.ac.uk} 
   \And
   Subramanian Ramamoorthy \\
   School of Informatics \\
   University of Edinburgh, UK \\
   \texttt{s.ramamoorthy@ed.ac.uk} 
   \And
   Jan Oliver Ringert \\
   Department of Computer Science \\
   Bauhaus University Weimar, Germany \\
   \texttt{jan.ringert@uni-weimar.de} 
   \And
   James Wilson \\
   Department of Engineering Mathematics \\
   University of Bristol, UK \\
   \texttt{j.wilson@bristol.ac.uk} 
   \And
   Shane Windsor \\
   Department of Aerospace Engineering \\
   University of Bristol, UK \\
   \texttt{shane.windsor@bristol.ac.uk} 
   \And
   Kerstin Eder \\
   Department of Computer Science \\
   University of Bristol, UK \\
   \texttt{kerstin.eder@bristol.ac.uk} 
}

\begin{document}
\maketitle
\begin{abstract}
As autonomous systems (AS) increasingly become part of our daily lives, ensuring their trustworthiness is crucial. In order to demonstrate the trustworthiness of an AS, we first need to specify what is required for an AS to be considered trustworthy. This roadmap paper identifies key challenges for specifying for trustworthiness in AS, as identified during the ``Specifying for Trustworthiness'' workshop held as part of the UK Research and Innovation (UKRI) Trustworthy Autonomous Systems (TAS) programme. We look across a range of AS domains with consideration of the resilience, trust, functionality, verifiability, security, and governance and regulation of AS and identify some of the key specification challenges in these domains. We then highlight the intellectual challenges that are involved with specifying for trustworthiness in AS that cut across domains and are exacerbated by the inherent uncertainty involved with the environments in which AS need to operate.
\end{abstract}

\section{Introduction}\label{introduction}
	Autonomous systems (AS) are systems that involve software applications, machines, and people, that is, systems that are capable of taking actions with little or no human supervision~\cite{Murukannaiah2020}. 
Soon, AS will no longer be confined to safety-controlled industrial settings. Instead, they will increasingly become part of our daily lives having matured across different domains like driverless cars, healthcare robotics, and uncrewed aerial vehicles (UAVs). As such, it is crucial that these systems are trusted and trustworthy. \emph{Trust} may vary, as it can be gained and lost over time. Different research disciplines define trust in different ways. This article focuses on the notion of trust that concerns the relationship between humans and AS. AS are considered \emph{trustworthy} when the design, engineering, and operation of these systems generate positive outcomes and mitigate outcomes which can be harmful ~\cite{Naiseh2022}.
The trustworthiness of AS can be dependent on many factors such as: (i) explainability, accountability, and understandability to different users; (ii) robustness of AS in dynamic and uncertain environments; (iii) assurance of their design and operation through verification and validation (V\&V) activities; (iv) confidence in their ability to adapt their functionality as required; (v) security against attacks on the systems, users, and deployed environment; (vi) governance and regulation of their design and operation; and (vii) consideration of ethics and human values in their deployment and use ~\cite{Naiseh2022}. 

There are various techniques for demonstrating the trustworthiness of systems, such as synthesis, formal verification at design time, runtime verification or monitoring, and test-based methods. 
However, common to all these techniques is the need to formulate \emph{specifications}. 
A specification is a detailed formulation that provides ``a definitive description of a system for the purpose of developing or validating the system''~\cite{ISO24765:2017}. 
According to~\cite{Kress-Gazit2021}, writing specifications that capture trust is challenging. A human will only trust an AS to perform in a safe manner (i.e. nothing bad happens), if it clearly and demonstrably acts in such a manner. This requires the AS to not only be safe, but also to be seen as safe by the human. In the same manner, it is equally important to ensure that the AS trusts the human~\cite{Kress-Gazit2021}. To address this, specifications need to go beyond typical functionality and safety aspects.

Engineering trustworthy and trusted AS involves different processes, technology, and skills than those required for traditional software solutions. Many practitioners in the AS or artificial intelligence (AI) domains have learnt by accumulating experiences, and failures, across projects~\cite{AmershiBBDGKNN019}. Best practices have started to emerge. There is increasing evidence of the need for rigorous specification techniques for developing and deploying AI applications~\cite{damour2020}. Even when not life critical, actions and decisions made by AS may have serious consequences. If we are to use them in our businesses, at doctor’s surgeries, on our roads, or in our homes, we need to build AS that precisely satisfy the requirements of their stakeholders. However, specifying requirements for AS (AI in particular) remains more a craft than it being a science, for e.g.\ machine learning (ML) applications are often specified based on optimisation and efficiency measures rather than well-specified quality requirements that relate to stakeholders needs~\cite{IshikawaM20} and further research is needed.		

In the UK Research and Innovation (UKRI) Trustworthy Autonomous Systems (TAS) programme, we conduct cross-disciplinary fundamental research to ensure that AS are safe, reliable, resilient, ethical, and trusted. TAS is organised around six research projects called Nodes and a Hub, and each Node focuses on the individual aspects of trust in AS, such as resilience, trust, functionality, verifiability, security, and governance and regulation. 

Undertaking a community approach, this roadmap paper is the result of the ``Specifying for Trustworthiness'' workshop held during the TAS All Hands Meeting in September 2021. The workshop gathered a diverse group of researchers from all parts of the TAS programme. 
Co-authored by a representative sample of the AS community in the UK, in this paper, we highlight the specification challenges for AS with illustrations from a representative set of domains currently being investigated within our community. 
The main contribution of this roadmap paper is to identify key open research problems termed `intellectual challenges' that are involved with specifying for trustworthiness in AS that cut across domains, and are exacerbated by the inherent uncertainty involved with the environments in which AS need to operate. 
We take a broad view of specification, concentrating on top-level requirements including but not limited to functionality, safety, security, and other non-functional properties that contribute to the trustworthiness of AS. 
Also, we intentionally leave the discussion on the formalisation of these specifications for the future, when the understanding of what is required to specify for trustworthiness is more mature. 

To motivate and present the research challenges associated with specifying for trustworthiness in AS, the rest of this paper is divided into three parts. Section~\ref{as-domains} discusses a number of AS domains, each with its unique specification challenges. Section~\ref{as-challenges} presents key intellectual challenges that are currently being investigated within our community. Finally, Section~\ref{conclusions} summarises our findings.

\section{Autonomous Systems Domains and their Specification Challenges}\label{as-domains}
	In this paper we classify AS domains based on two criteria: the number of autonomous agents (single or multiple), and whether humans are interacting with the AS as part of the system or the environment, following~\cite{Schneiders2022}. Accordingly, we distinguish AS domains into four categories: (i) a single autonomous agent (e.g. automated driving, UAV); (ii) a group of autonomous agents (e.g. swarms); (iii) an autonomous agent assisting a human (e.g. AI in healthcare, human–robot interaction); and (iv) a group of autonomous agents collaborating with humans (e.g. emergency situations, disaster relief). We discuss the specification challenges involved with AS using illustrations from a representative set of domains, as being investigated within our community in TAS (see Table~\ref{AS-challenges}), rather than attempting to cover all possible AS domains. 
\begin{itemize}[leftmargin=0.5cm]
	\item \textbf{Single Autonomous Agent: Automated Driving, UAV} 
\end{itemize}
Automated driving (self-driving) refers to a class of AS that vary in the extent to which they make decisions independently (SAE J3016 standard taxonomy). The higher levels of autonomy, levels 3-5, refer to functionality ranging from traffic jam chauffeur to completely hands-free driving in all conditions. Despite an explosion of activity in this domain in recent years, the majority of systems being considered for deployments depend on careful delineation of the operation design domain to make the specification of appropriate behaviour tractable. Even so, the specification problem remains difficult for a number of reasons. 
Firstly, traffic regulations are written in natural language, ready for human interpretation. Although the highway code rules are intended for legal enforcement, they are not specifications that are suitable for machines. 
There typically are many exceptions, context-dependent conflicting rules and guidance of an `open nature', all of these require interpretation in context. 
Driving rules can often be vague or even conflicting and may need a base of knowledge to interpret the rule given a specific context. The UK Highway Code Rule 163 states that after you have started an overtaking manoeuvre you should ``move back to the left as soon as you can but do not cut in''~\cite{UKHighwayCode22}. A more explicit specification of driving conduct (e.g. Rule 163) to something more machine interpretable that captures the appropriate behaviour presents a challenge to this research area. When people are taught to perform this activity, a significant portion of the time is spent in elaborating these special cases, and much of the testing in the licensing regime is aimed at probing for uniformity of interpretation. How best to translate these human processes into the AS domain is important not only for achieving safety but also acceptability.
Secondly, driving in urban environments is an intrinsically interactive activity, involving several actors whose internal states may be opaque to the automated vehicle. As an example, the UK Highway Code asks drivers to not ``pull out into traffic so as to cause another driver to slow down''. 
Without further constraint on what the other drivers could possibly do, specifying appropriate behaviour becomes difficult, and any assumptions made in that process would call into question the safety of the overall system when those assumptions are violated. 
Thus, two key challenges in the area of automated driving are the lack of machine-readable specifications that formally express acceptable driving behaviour and the need to specify the actions of other road users (see also Table~\ref{AS-challenges}). To some extent, these issues arise in all open environments. However, in automated driving, the task is so intricately coupled with the other actors that even the default assumptions may not be entirely clear, and the relative variation in behaviour due to different modelling assumptions could be qualitatively significant.

A UAV or drone is a type of aerial vehicle that is capable of autonomous flight without a pilot on board. UAVs are increasingly being applied in diverse applications, such as logistics services, agriculture, emergency response, and security. Specification of the operational environment of UAVs is often challenging due to the complexity and uncertainty of the environments that UAVs need to operate in. 
For instance, in parcel delivery using UAVs in urban environments, there can be uncertain flight conditions (e.g.\ wind gradients), and highly dynamic and uncertain airspace (e.g.\ other UAVs in operation). 
Recent advances in machine learning offer the potential to increase the autonomy of UAVs in uncertain environments by allowing them to learn from experience. 
For example, machine learning can be used to enable UAVs to learn novel manoeuvres to achieve perched landings in uncertain windy conditions~\cite{Fletcher2021}. 
In these contexts, a key challenge is how to specify how the system should deal with situations that go beyond the limits of its training (Table~\ref{AS-challenges}).

\footnotesize
\begin{table}  
	\caption{AS domains and their specification challenges.} \label{AS-challenges}   

	\begin{tabular}[width=\textwidth]{ |p{3cm}|p{3cm}|p{10cm}| } 
		\hline
		\hfil \textbf{Category} & \hfil \textbf{Domain} &  \hfil \textbf{Specification Challenge} \\
		\hline
		\multirow{2}{3cm}{\vfil Single \newline Autonomous Agent} & \multirow{3}{3cm}{Automated Driving} & How to address the lack of machine-readable specifications that formally express acceptable driving behaviour? \\ \cline{3-3} & & How to specify the actions of other road users? \\ \cline{2-3}
		& \vfil UAV & How to specify how the UAV should deal with situations that go beyond the limits of its training? \\ 
		\hline
		{\vfil Multiple \newline Autonomous Agents} & \vfil Swarms & How to specify the emergent behaviour of a swarm that is a consequence of the interaction of individual agents with each other and the environment? \\
		\hline
		\multirow{6}{3cm}{Autonomous Agent Assisting a Human}
		& \multirow{3}{3cm}{Human--\newline Robot Interaction} & How to specify the perceptual, reasoning, and behavioural processes of robot systems? \\ \cline{3-3} & & How to infer human mental states interactively? \\ \cline{2-3}
		& \multirow{3}{3cm}{\vfil AI in Healthcare} & How to specify `black box' models? \\ \cline{3-3} & & What is the role of explainability and faithfulness of the interpretation of semantics? \\ \cline{3-3}& & What is the role of pre-trained models in pipelines? \\ \cline{2-3}
		\hline
		{\vfil Multiple \newline Autonomous Agents \newline Collaborating with Humans}  & \multirow{2}{3cm}{\vfill Emergency Situations \& Disaster Relief} & {\vfill How to specify collaboration between autonomous agents and different human agents in emergency settings?} \\ \cline{3-3} &  & How to specify security where large amounts of data need to be collected, shared, and stored? \\ 
		\hline
	\end{tabular}
\end{table}
\normalsize	
\begin{itemize}[leftmargin=0.5cm]
	\item \textbf{Multiple Autonomous Agents: Swarm Robotics}
\end{itemize}
Swarm robotics provides an approach to the coordination of large numbers of robots, which is inspired from the observation of social insects \cite{Sahin2005}. Three desirable properties in any swarm robotics system are robustness, flexibility and scalability. 
The functionality of a swarm is emergent (e.g. aggregation, coherent ad hoc network, taxis, obstacle avoidance and object encapsulation \cite{Winfield2006}), and evolves based on the capabilities of the robots and the numbers of robots used. 
The overall behaviours of a swarm are not explicitly engineered in the system, as they might be in a collection of robots that are all centrally controlled, but they are an emergent consequence of the interaction of individual agents with each other and the environment.
This emergent functionality poses a challenge for specification. The properties of individual robots can be specified in a conventional manner, yet it is the emergent behaviours of the swarm that determine the performance of the system as a whole. The challenge is to develop specification approaches that specify properties at the swarm level that can be used to develop, verify and monitor swarm robotic systems.
\begin{itemize}[leftmargin=0.5cm]
	\item \textbf{Autonomous Agent Assisting a Human: Human–Robot Interaction, AI in Healthcare}
\end{itemize}
Interactive robot systems aim to complete their tasks while explicitly considering states, goals and intentions of the human agents they collaborate with, and aiming to calibrate the trust humans have for them to an appropriate level. This form of human-in-the-loop real-time interaction is required in several application domains including assistive robotics for activities of daily living \cite{GaoEtAl2020}, healthcare robotics, shared control of smart mobility devices \cite{SohDemiris2015}, and collaborative manufacturing among others. Most specification challenges arise from the need to provide specifications for the perceptual, reasoning and behavioural processes of robot systems that will need to acquire models of, and deal with, the high variability exhibited in human behaviour. 
While several human-in-the-loop systems employ mental state inference, the necessity for interactively performing such inference (including as beliefs and intentions), typically through sparse and/or sensor data from multimodal interfaces, imposes further challenges for the principled specification of human factors and data-driven adaptation processes in robots operating in close proximity to humans, where safety and reliability are of particular importance. 

Healthcare is a broad application domain which already enjoys the many benefits arising from the use of AI and AI-enabled autonomy. This has ranged from more accurate and automated diagnostics, to a greater degree of autonomy in robot surgery, and entirely new approaches to drug discovery and design. The use of AI in medical diagnosis has advanced to an extent that in some settings, e.g. mammography screening, automated interpretation seems to match human expert performance in some trials. However, there remains a gap in test accuracy. It has been argued that the automated systems are not sufficiently specific to replace radiologist double reading in screening programmes~\cite{Freemann1872}. These gaps also highlight the main specification challenges in this domain. Historically, the human expertise in this domain has not been explicitly codified, so that it can be hard to enumerate desired characteristics. It is clear that the specifications must include notions of invariance to instrument and operator variations, coverage of condition and severity level, etc. Beyond that, the `semantics' of the biological features used to make fine determinations are subject to both ambiguity or informality, and variability across experts and systems. Moreover, the use of deep learning to achieve automated interpretation brings with it the need for explainability. This manifests itself in the challenge of guarding against `shortcuts'~\cite{degrave2021ai}, wherein the AI diagnostic system achieves high accuracy by exploiting irrelevant side variables instead of identifying the primary problem (e.g. radiographic COVID-19 detection using AI~\cite{degrave2021ai}). 
The specific challenge here is how to specify with respect to `black box' models. In this regard, we can highlight the role of explainability and faithfulness of interpretation of semantics, and the role of pre-trained models in pipelines (see Table~\ref{AS-challenges}). 

\begin{itemize}[leftmargin=0.5cm]
	\item \textbf{Multiple Autonomous Agents Collaborating with Humans: Emergency Situations, Disaster Relief}
\end{itemize}
Emergency situations evolve dynamically and can differ in terms of the type of incident, its magnitude, additional hazards, the number and location of injured people. They are also characterised by urgency; they require a response in the shortest time frame possible and call for a coordinated response of emergency services and supporting organisations, which are increasingly making use of AS. This means that successful resolutions depend not only on effective collaboration between humans~\cite{james2011organizational}, but also between humans and AS. Thus, there is a need to specify both functional requirements and SLEEC requirements (the Social, Legal, Ethical, Empathic, Cultural rules and norms that govern an emergency scenario). AS in emergency response contexts vary hugely, and as such, the kinds of SLEEC issues pertaining to them need to be incorporated into the design process, rather than implemented afterwards. This suggests a shift from a static design challenge towards the need to specify for adaptation to the diversity of emergency actors and complexity of emergency contexts, which are time-sensitive, and involve states of exception not common in other open AS environments such as autonomous vehicles. In addition, to enhance collaboration between autonomous agents and different human agents in emergencies, specifying human behaviour remains one of the main challenges in emergency settings. 

There are also challenges for specifying security in the context of disaster relief. A large part of this comes from the vast amounts of data that needs to be collected, shared and stored between different agencies and individuals. Securing a collaborative information management system is divided between technical forms of security, such as firewalling and encryption, and social forms of security, such as trust. To provide security to a system, both aspects must be addressed in relation to each other within a specification. 	

\section{Intellectual Challenges for the Autonomous Systems Community}\label{as-challenges}
The preceding section discussed specification challenges that are unique to a representative set of domains investigated within our community. Now we discuss ten `intellectual challenges' that are involved with specifying for trustworthiness in AS that can cut across domains, and are exacerbated by the inherent uncertainty involved with the environments in which AS need to operate. These challenges were identified out of stimulating discussions among the speakers and participants of the breakout groups at the ``Specifying for Trustworthiness'' workshop. 

\begin{figure*}
	\centering
	\includegraphics[width=1.0\textwidth]{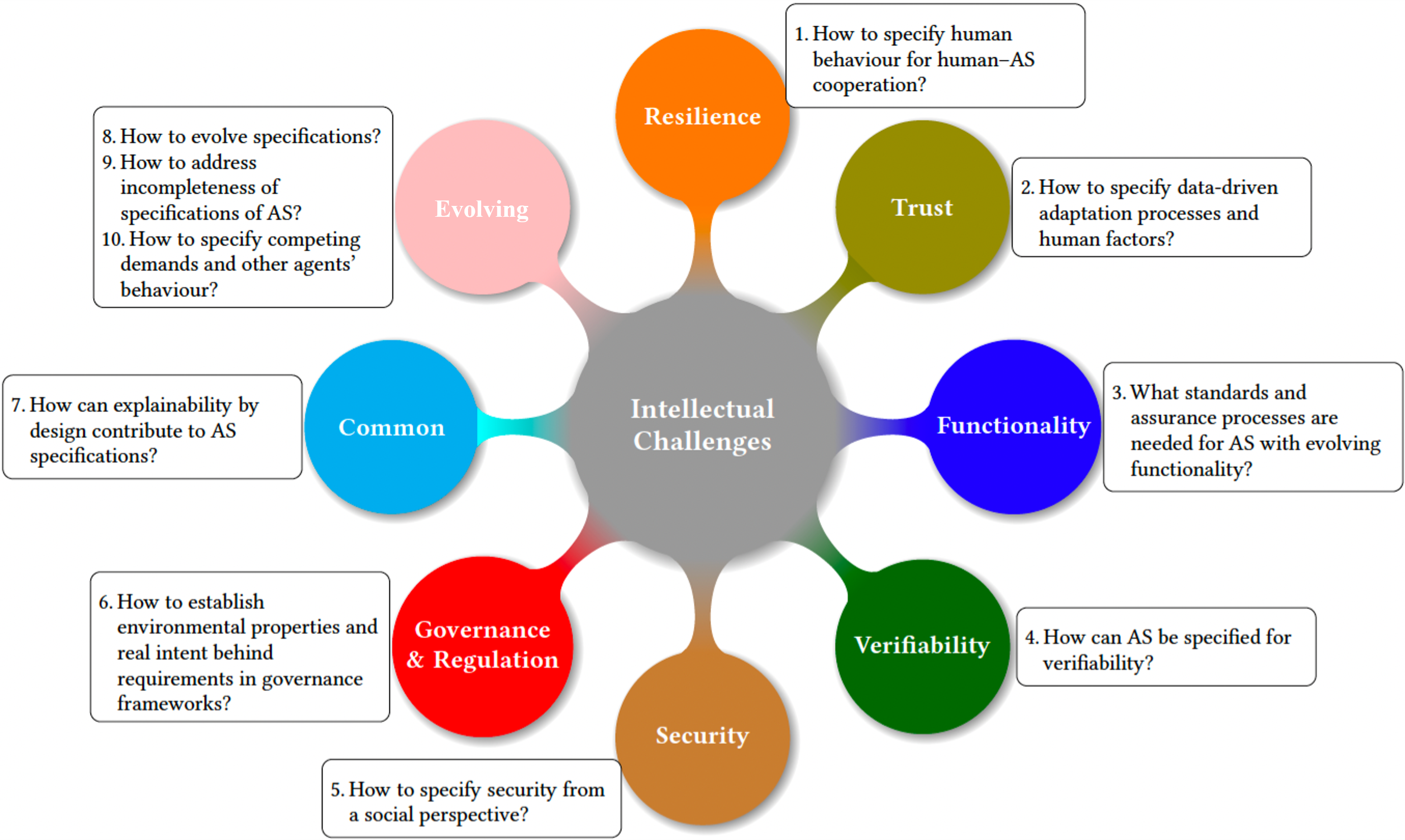}
	\caption{Intellectual challenges for the AS community.}	
	\label{Intellectual-challenges}
\end{figure*}

The intellectual challenges 1--6 are in the six \emph{focus areas} of trust in AS (i.e. resilience, trust, functionality, verifiability, security, and governance and regulation), as identified by their respective speakers.
Meanwhile, the remaining four challenges have either a \emph{common} focus (7) across the TAS programme, or they are \emph{evolving} in nature (8--10) (see Fig.~\ref{Intellectual-challenges}). 
For each challenge we provide an overview, identify high-priority research questions, and suggest future directions.

Many of the specification challenges to be discussed are shared by systems such as multi-agents systems, cyber-physical-social systems, or AI-based systems. Autonomy is an important characteristic of these systems, and so is the need for trustworthiness. Specification challenges have also received a lot of attention in ‘non-AS’, e.g. safety-critical systems. Yet, many of the challenges are exacerbated in AS because of the inherent uncertainty of their operating environment: they are long-lived, continuously running systems that interact with the environment and humans in ways that can hardly be fully anticipated at design time and continuously evolve at runtime. In other words, while those challenges are not specific to AS, AS exacerbate them. 

\begin{enumerate}[leftmargin=0.5cm]
	
	\item \textbf{How to specify human behaviour for human--AS cooperation?}
\end{enumerate}
How to model human behaviour to enable cooperation with AS is challenging but crucial for the resilience of the system as a whole. It is the diversity in human enactment that drives the uncertainty about what people do and don't do, and, subsequently, the way human behaviour can be specified. Knowing the mental state of others enables AS to steer a cooperation that is consistent with the needs of the AS, as well as to respond to the needs of human agents in an appropriate manner. 

Different theories of human behaviours explain diversity in human action in different ways and by detecting various determinants of human behaviour. For example, a behaviourist approach suggests that every behaviour is a response to a certain stimulus~\cite{heimlich2008understanding}. Albeit true, this approach is restrictive in addressing the complexity of human behaviour, as well as the different ways that human behaviour develops during cooperation. To grasp that humans are embodied with purposes and goals that affect each other, the concept of joint-action can be introduced as ``a social interaction whereby two or more individuals coordinate their actions in space and time to bring about change in the environment''~\cite{sebanz2006joint}. Adapting it to human–robot interaction, this approach is suggestive of an interplay between humans and AS, such that what matters is not only how the AS understands the system but also how humans understand the way the autonomous agent behaves and is willing to cooperate~\cite{grigore:2013}. Thus, cooperation arises from a shared understanding between agents, which is a challenge to specify. 

The social identity approach~\cite{spears2021social} induces this concept of a shared understanding by providing an explanation of human behaviour focusing on how social structures act upon cognition. It proposes that, alongside our personal identity, our personality - who we are, we also have multiple social identities based on social categories and groups. 
Previous research has shown that social identities influence people's relation with technology~\cite{lee2001effect}. Sharing a social identity initiates pro-social behaviours, such as helping behaviours in emergency situations~\cite{drury2018role}. People adapt their behaviour in line with their shared identities, which in turn, enhances resilience. Specifying social identities to enable cooperation is challenging. It requires answering questions such as: how do we represent different identities and how do we reason about them? Following the social identity approach to specify identities for human-autonomous agent cooperation requires an investigation of how to operationalise social identity, a psychological state, into software embedded within AS.

\begin{enumerate}[leftmargin=0.5cm]
	\setcounter{enumi}{1}
	\item \textbf{How to specify data-driven adaptation processes and human factors?}
\end{enumerate}
Specifying, designing, implementing, and deploying interactive robot systems that are trustworthy for use in scenarios where humans and robots collaborate in close proximity is challenging, given that safety and reliability in such scenarios are of particular importance. Example scenarios include assisting people with activities of daily living such as mobility~\cite{SohDemiris2015} and dressing~\cite{GaoEtAl2020}, rehabilitation robotics, adaptive assistance in intelligent vehicles, robot assistants in care homes and hospital environments, among others. The intellectual challenge the AS community faces is the specification, design and implementation of trustworthy perceptual, cognitive, and behaviour generation processes that explicitly incorporate parametrizable models of human skills, beliefs, and intentions~\cite{Demiris2007}. 
These models are necessary for interactive assistive systems since they need to decide not only how but also when to assist~\cite{GeorgiouDemiris}. Given the large variability of human behaviour, the parameters of these user models need to be acquired interactively, typically from sparse and potentially noisy sensor data, a particularly challenging inverse problem. An additional challenge is introduced in the case of long-term human–robot interaction, where the assistive system needs to learn and take into consideration human developmental aspects, typically manifested in computational learning terms as model drift. 
As an example, consider an assistive mobility device for children with disabilities~\cite{SohDemiris2015}: as the child's perceptual, cognitive, emotional and motor skills develop over time, their requirements for the type, amount and frequency of the provided assistance will need to evolve. Similarly, when assisting an elderly person or someone recovering from surgery, the distributions of the human data that the robot sensors collect will vary not only according to the context but also over time. 
Depending on the human participant, and their underlying time-varying physiological and behavioural particularities, model drift can be sudden, gradual, or recurring, posing significant challenges to the underlying modelling methods. Principled methods for incorporating long-term human factors into the specification, design and implementation of assistive systems, that adapt and personalise their behaviour for the benefit of their human collaborator, remain an open research challenge. 
\begin{enumerate}[leftmargin=0.5cm]
	\setcounter{enumi}{2}
	\item \textbf{What standards and assurance processes are needed for AS with evolving functionality?}
\end{enumerate}
AS with \emph{evolving functionality}---the ability to adapt and change in function over time---pose significant challenges to current processes for specifying functionality. 
Most conventional processes for defining system requirements assume that these are fixed and can be defined in a complete and precise manner before the system goes into operation. 
Existing standards and regulations do not accommodate the adaptive nature of AS with evolving functionality. This is a key limitation~\cite{Fisher2020} that prevents promising applications such as swarm robots which adapt through emergent behaviour and UAVs with ML-based flight control systems from deployment. 

For airborne systems and in particular for UAVs, several industry standards and regulations have been introduced to specify requirements for system design and safe operation (e.g.\ DO-178C, DO-254, ED279, ARP4761, NATO STANAG 4671, CAP 722). However, none of these standards or regulations cover the types of ML-based systems which are currently being developed to enable UAVs to operate autonomously in uncertain environments.

The ability to adapt and to learn from experience are important abilities to enable AS to operate in real world environments. When one considers the existing industry standards, they are either implicitly or explicitly based on the V\&V model, which moves from requirements through design onto implementation and testing before deployment~\cite{Jia2021}. 
However, this model is unlikely to be suitable for systems with the ability to adapt their functionality in operation; e.g.\ through interaction with other agents and the environment, as is the case with swarms; or through experience-driven adaptation as is the case with machine learning. AS with evolving functionality follow a different, much more iterative life-cycle. 
Thus, there is a need for new standards and assurance processes that extend beyond design time and allow continuous certification at runtime. 

\begin{enumerate}[leftmargin=0.5cm]
	\setcounter{enumi}{3}
	\item \textbf{How can AS be specified for verifiability?}
\end{enumerate}
For a system to be {\em verifiable\/}, a person or a tool needs to be able to check its correctness~\cite{ISO24765:2017} with respect to its requirements and specification. 
The main challenge is in specifying and designing the system in such a way that this process is made as easy and intuitive as possible.
For AS in particular, specific challenges include 
(i) capturing and formalising requirements including functionality, safety, security, performance and, beyond these, any additional non-functional requirements purely needed to demonstrate trustworthiness; 
(ii) handling flexibility, adaptation and learning; and 
(iii) managing the inherent complexity and heterogeneity of both the AS and the environment it operates in. 

Specifications need to represent the different aspects of the overall system in a way that is natural to domain experts, facilitates modelling and analysis, provides transparency of how the AS works and gives insights into the reasons that motivate its decisions. 
To specify for verifiability, a specification framework will need to offer a variety of domain abstractions to represent the diverse, flexible and possibly evolving requirements AS are expected to satisfy. 
Furthermore, the underlying verification framework should connect all these domain abstractions to allow an analysis of their interaction. This is a key challenge in specification for verifiability in AS.

AS can be distinguished using two criteria: the degree of autonomy and adaption, and the criticality of the application which can range from harmless to safety-critical. 
We can consider which techniques or their combinations are needed for V\&V at the different stages of the system life-cycle. 
When AS operate in uncontrolled environments, where there is a need for autonomy, learning and adaptation, the need for runtime V\&V emerges. 
There, a significant challenge is finding rigorous techniques for the specification and V\&V of safety-critical AS where requirements are often vague, flexible and may contain uncertainty and fuzziness. 
V\&V at design time can only provide a partial solution, and more research is needed to understand how best to specify and verify learning and adaptive systems by combining design time with runtime techniques. 
Finally, identifying the design principles that enable V\&V of AS is a key pre-requisite to promote verifiability to a first-class design goal alongside functionality, safety, security, and performance.
\begin{enumerate}[leftmargin=0.5cm]
	\setcounter{enumi}{4}
	\item \noindent\textbf{How to specify security from a social perspective?} 
\end{enumerate}
There are technical sides to security, but there are also social dimensions that matter when considering how an AS enforces its status as secure. In this context, security overlaps with trust. One can only be assured a system is secure, if one trusts that system. Public trust is a complex issue, shot through with media, emotions, politics, and competing interests. How do we go about specifying security in a social sense? 

On the technical side, there are fairly specific definitions for specification which can be grasped and measured. From the social perspective, the possibility of specification relies on a network of shared assumptions and beliefs that are difficult to unify. In fact, much of the value from engagement over social specifications derives from the diversity and difference. 
A predominant concern in social aspects of security is where data is shared between systems (social-material interactions).
That is, whenever an AS communicates with a human being or an aspect of the environment. 
Although these interactions have technical answers, to find answers that consider social science perspectives requires collaboration and agile methods to facilitate that collaboration.

The human dimension means that it is not enough to specify technical components. Specifications must also capture believes, desires, fears and at times misinformation with respect to how those are understood, regarded and perceived by the public. For example, in what ways can we regard pedestrians as passive users of automated vehicles? How are automated vehicles regarded by the public, and how are pedestrians involved in automated mobility?

The ethical challenges that emerge for AS security also relate to the legal and social ones. 
The difficulty centres around how to create regulations and specifications on a technical level, that are also useful socially, facilitating responsiveness to new technologies that are neither simply techno-phobic nor passively accepting. 
Doing so must involve both innovation and public input, so that the technology developed works for everyone. The ELSI (Ethical, Legal \& Social Implications) framework~\cite{Escalante2019} is an example of a framework aimed to engage designers, engineers, and public bodies in answering these questions. ELSI is an inherently cross-disciplinary set of approaches for tackling AS security, as many interrelated and entangled aspects. Specifying security requires connection, collaboration, and agile ethical methods.	
\begin{enumerate}[leftmargin=0.5cm]
	\setcounter{enumi}{5}
	\item \textbf{How to establish environmental properties and real intent behind requirements in governance frameworks?}
\end{enumerate}
Computer scientists treat specifications as precise objects, often derived from requirements by purging features such that they are defined with respect to environment properties that can be relied on regardless of the machine's behaviour. Emerging AS applications in human-centred environments can challenge this way of thinking, particularly so because the environment properties may not be fully understood, or because it is hard to establish if the real intent behind a requirement can be verified. These gaps should be addressed in governance frameworks in order to engender trust.

For instance, in all of the domains mentioned, we are increasingly seeing systems that are data-first and subject to continuous deployment. This has the interesting consequence that sometimes even the task requirements are really only given in terms of fit to observed human behaviour~\cite{topol2019high}, e.g.\ what does an expert say in radiology interpretation and does the AI-based AS match that~\cite{Freemann1872}. We see this as a crucial area for future development, as existing workflows depend on human interpretation of rules in crucial ways, whereas when AS perform the same decisions there is scope for significant disruption of these workflows due to potential gaps that become exposed.

Furthermore, many emerging concerns such as fairness are not only difficult to formalise in the sense of software specification, but also their many definitions can be conflicting such that it is impossible to satisfy all of them in a given system~\cite{narayanan21fairness}. 

AS of the future will need a combination of informal and formal mechanisms for governance. In domains such as automated vehicles, trustworthiness of the system may require a complete ecosystem approach~\cite{koopman2019certification} involving community-defined scenario libraries, enabling the greater use of simulation in verification, and independent audits via independent third parties. This calls for the development of new computational tools for performance and error characterisation, systematic adversarial testing with respect to a range of different specification types, and causal explanations that address not only a single instance of a decision but better expose informational dependencies that are useful for identifying edge cases and delineating operational design domains. 

In addition to all of these technical tools, there is a need to understand the human-machine context in a more holistic manner, as this is really the target of effective governance. People's trust in an AS is not determined by technical reliability alone. Instead, the expectations of responsibility and accountability are associated with the human team involved in the design and deployment of the system, and the organisational design behind the system. A vast majority of system failures are really failures arising from mistakes made in this `outer loop'. Therefore, effective regulations must first begin with a comprehensive mapping of responsibilities that must be governed, so that computational solutions can be tailored to address these needs. Furthermore, there is a need for ethnographic understanding of AS being used in context, which could help focus technical effort on the real barriers to trustworthiness.
\begin{enumerate}[leftmargin=0.5cm]
	\setcounter{enumi}{6}
	\item \textbf{How can explainability by design contribute to AS specifications?}
\end{enumerate} 	
There are increasing calls for explainability in AS, with emerging frameworks and guidance~\cite{Hamon:2020} pointing to the need for AI to provide explanations about decision making. A challenge with specifying such explainability is that existing frameworks and guidance are not prescriptive: what is an actual explanation and how should an explanation be constructed? Furthermore, frameworks and guidance tend to be concerned with AI in general, and not AS.

A case study addressing regulatory requirements on explainability of automated decisions in the context of a loan application~\cite{Huynh:DGOV21} provided foundations for a systematic approach. Within this context, explanations can act as external detective controls, as they provide specific information to justify the decision reached and help the user take corrective actions~\cite{Tsakalakis:CLSR21}. But explanations can also act as internal detective controls, i.e. a mechanism for organisations to demonstrate compliance to the regulatory frameworks they have to implement. 
The study and design of AS include many facets: not only black-box or grey-box AI systems, but also the various software and hardware components of the system, the curation and cleansing of datasets used for training and validation, the governance of such systems, their user interface and crucially the users of such systems with a view of ensuring that they do not harm but provide benefits to these users and society in general. There are typically a range of stakeholders involved, from the designers of the systems, to their hosts and/or owners, their users (consumers and operators), third-parties and increasingly regulators. In this context, many questions related to trustworthy AS (such as what is an actual explanation and how should an explanation be constructed? What is the purpose of an explanation? What is the audience of an explanation? What is the information that it should contain?~\cite{Huynh:DGOV21,Tsakalakis:CLSR21}) have to be addressed holistically. It no longer suffices to focus on the explainability of a black-box decision system, its behaviour needs to be explained, with more and less details, in the context of the overall AS. However, to adequately address these questions, explainability should not be seen as an after-thought, but as an integral part of the specification and design of a system, leading to explainability requirements to be given the same level of importance as all other aspects of a system. 

In the context of trustworthy AS, emerging AS regulations could be used to drive the socio-technical analysis of explainability. A particular emphasis would have to be on the autonomy and the hand-over between systems and humans that characterise trustworthy AS. The audience of explanations will also be critical, from users and consumers to businesses, organisations and regulators. Finally, considerations for post-mortem explanations, in case of crash or disaster situations involving AS, should lead to adequate architectural design for explainability.	
\begin{enumerate}[leftmargin=0.5cm]
	\setcounter{enumi}{7}
	\item \textbf{How to evolve specifications?}
\end{enumerate}
Every typical AS undergoes changes over its lifetime that require going beyond an initially specified spectrum of operation --- despite the observation that this spectrum is typically quite large for AS in the first place. The evolution of trustworthy AS may concern changes of the requirements of their functional or non-functional properties, changes of the environment that the AS operate in, and changes in the trust of users and third parties towards the AS. 

Initial specifications of the AS may no longer reflect the desired properties of the system or they may fail to accurately represent its environment. The evolution of specifications presents challenges in balancing the autonomy of the system.

While any non-trivial system requires evolution and maintenance~\cite{Mens08}, some challenges are exacerbated for trustworthy AS. As an example, observed changes in trust towards the AS might require changes to behaviour specifications, even if the AS operations are perfectly safe. Conversely, required changes to specifications might have negative impacts on future trust towards the AS. New methods will be required to efficiently deal with the various dimensions of trust in evolution of specifications.

One dimension of trust relates to transparency towards developers of AS specifications. Approaches that compare evolving specifications on a syntactical level as currently done for code, or based on metrics as currently done for AI models, are unlikely to be sufficient for effective maintenance. Analysis will need to scale beyond syntactic differences to include also semantic differences~\cite{MaozR18} and allow for efficient analysis of the impact of changes on the level of systems rather than artefacts. New techniques to compare specifications of AS are required that identify, present, and explain differences as well as their potential impact on the system's trustworthiness.
\begin{enumerate}[leftmargin=0.5cm]
	\setcounter{enumi}{8}
	\item \textbf{How to address incompleteness of specifications of AS?}
\end{enumerate}
Incompleteness is a common property of specifications. Only the use of suitable abstractions allows for coping with the complexity of systems~\cite{Kramer07}. However, there is an important difference in the incompleteness introduced by abstractions, the process of eliminating unnecessary detail to focus, e.g.\ on behavioural, structural, or security-related aspects of a system, and the incompleteness related to the purpose of the specification, i.e.\ the faithful representation of the system in an abstraction.

On the one hand, if the purpose of creating and analysing a specification is to examine an AS and to learn about possible constraints, then incompleteness of the AS representation in the specification is important as it allows for obtaining feedback with low investment in specification development~\cite{Jackson19}, e.g. for the reduction of ambiguities. 
On the other hand, if the purpose of the specification is to prove a property, then incompleteness of the AS representation may lead to incorrect analyses results manifesting in false positives or false negatives. False positives are often treated by adding the missing knowledge to the specification of AS. For example, verifying the specification of an infusion pump reported a false positive due to incompleteness~\cite{HarrisonMCC17}. The specification had to be changed to a ``much more complex''~\cite{HarrisonMCC17} one to remove the false positive.

One way of addressing incompleteness are partial models~\cite{WeiGC11,FamelisSC12,FischbeinDBCU12} where models and analyses are extended with modalities qualifying their completeness. The various approaches provide analysis of either syntactic properties~\cite{FamelisSC12} or behaviour refinements~\cite{WeiGC11,FischbeinDBCU12}. Combinations and extensions to rich specification languages for AS are part of this research challenge.

In addition to analysis tasks, specifications are also used in synthesis tasks, and this is where the incompleteness of AS specifications can manifest itself in the construction of biased or incorrect systems. 
As an example, consider the specification of a robot operating in a warehouse~\cite{MaozR18robot}. The specification requires that the robot never hits a wall. With no assumptions about the environment, the synthesiser would take the worst-case view, i.e.\ walls move and hit the robot, and consequently report that the specification is not realisable and no implementation exists. Adding the assumption that walls cannot move as an environment constraint changes the outcome of the synthesis. 
Interestingly, when formulating requirements for humans, common sense allows us to cope with this type of incompleteness. 
However, the automated analysis of specifications for AS brings with it the challenge of identifying and handling (all) areas of incompleteness.

\begin{enumerate}[leftmargin=0.5cm]
	\setcounter{enumi}{9}
	\item \textbf{How to specify competing demands and other agents' behaviour?}
\end{enumerate}
Conventional approaches to V\&V for AS may seek to attain coverage against a specification to demonstrate assurance of functionality and compliance with safety regulations or legal frameworks. 
Such properties may be derived from existing legal or regulatory frameworks, e.g.\ the UK Highway Code for driving, which can then be converted into formal expressions for automatic checking~\cite{harper2021safety}.

But optimal safety does not imply optimal trust, and just because an AS follows rules does not mean that it will be accepted as a trustworthy system in human society. Other factors of trustworthiness should be considered, such as reliability, robustness, cooperation, and performance. 
We can also say that strictly following safety rules may even be detrimental to other trustworthiness properties, e.g.\ performance. 
Consider an automated vehicle trying to make progress through a busy market square full of people slowly walking across the road uncommitted to the usual observation of road conduct. 
The \emph{safest} option for the AS is to wait until the route ahead is completely clear before moving on, as by taking this option you do not endanger any other road user. However, \emph{better performance} may be to creep forward in a bid to promote your likelihood of success.
Driving then, is much more than following safety rules, which makes this a particularly hard specification challenge. In this scenario an assertive driving style would make more progress than a risk-averse one. 

In reality there will be significantly more considerations than just safety and performance, but this example illustrates the principle of conflicting demands between assessment standards. 
Consideration of other agents, such as properties of fairness or cooperation, would lead to a more trustworthy system. 
Additionally, the interaction of AS with people may require insight into \emph{social norms} of which there is no written standard by which these can be judged. Will the task of specification first require a codex of social interaction norms to be drawn together to add to the standards by which trust can be measured? 
Specifications would need to be written with reference to these standards, regulations and ethical principles, some of which do not currently exist, to ensure that any assessment captures the full spectrum of these trustworthiness criteria.

\section{Summary and Conclusions}\label{conclusions}	
As AS are becoming part of our daily lives and interacting more closely with humans, we need to build systems worthy of trust regarding safety, security, and other non-functional properties. In this paper we have first examined AS domains of different levels of maturity and then identified their specification challenges and related research directions. 
One of these challenges is the formalisation of knowledge easily grasped by humans so that it becomes interpretable by machines. Prominent examples include the specification of driving regulations for AVs, and the specification of human knowledge expertise in the context of AI-based medical diagnostics. 
How to specify and model human behaviour, intent and mental state is a further challenge that is common to all domains where humans interact closely with AS, such as in human-robot collaborative environments as found in smart manufacturing. 
Alternative approaches involve the specification of norms to characterise the desired behaviour of AS, which regulate what the system should do or should not do. An emerging direction of research is the design of monitors to observe the system and check compliance with norms~\cite{Criado:2018}. 
%
The example of swarm robotics raises the need and challenge to specify behaviour that emerges at the system level and relies on certain actions of the entities that form the system with each other and their environment. 

Beyond the technical aspects, across the specific AS domains, are research challenges related to governance and regulation for trustworthiness, requiring a holistic and human centred approach to specification focused on responsibility and accountability, and enabling explainability from the outset. 
Fundamental to specifying for trustworthiness is a sound understanding of human behaviour and expectations, as well as the social and ethical norms applicable when humans directly interact with AS. 
As for future work, an interesting extension of this paper would be to produce a classification of properties to be specified for trustworthiness under the different intellectual challenges discussed (e.g. socio-technical properties of explainability are purpose, audience, content, timing and delivery mechanism of explanations). 

We conclude that specifying for trustworthiness requires advances on the technical and engineering side, informed by new insights from social sciences and humanities research. Thus, tackling this specification challenge necessitates tight collaboration of engineers, roboticists and computer scientists with experts from psychology, sociology, law, politics and economics, as well as ethics and philosophy. Most importantly, continuous engagement with regulators and the general public will be key to trustworthy AS.

\section*{Acknowledgments}
This work has been supported by the UK EPSRC under the grants: [EP/V026518/1], [EP/V026607/1], [EP/V026747/1], [EP/V026763/1], [EP/V026682/1], [EP/V026801/2] and [EP/S027238/1]. Y.D. is also supported by a RAEng Chair in Emerging Technologies [CiET1718\textbackslash46].

\bibliographystyle{unsrt}  
\bibliography{TAS-Spec-arXiv}  

\end{document}